\documentclass[a4paper, UKenglish, cleveref, autoref, thm-restate]{lipics-v2021}
\pdfoutput=1
\hideLIPIcs

\graphicspath{{figs/}}

\title{A Study of Parallel Continuous Local Search}

\author{Cody J. Christopher}{School of Computing, Australian National University, Canberra, Australia.}{cody.christopher@anu.edu.au}{https://orcid.org/0000-0001-8444-2292}{}

\author{Charles Gretton}{School of Computing, Australian National University, Canberra, Australia.}{charles.gretton@anu.edu.au}{https://orcid.org/0000-0001-9803-0168}{}

\authorrunning{C.\,J. Christopher and C. Gretton}

\Copyright{Cody J Christopher and Charles Gretton}



\supplement{\afs{} implementation with \texttt{Apache-2.0}/\texttt{GPL-2.0-or-later} licenses:}
\supplementdetails[linktext={},
                   cite={},
                   subcategory={Source Code}, 
                   swhid={}
                   ]{Software}{https://github.com/cjchristopher/accelerated-fourier-sat/}

\nolinenumbers

\EventEditors{}
\EventNoEds{0}
\EventLongTitle{}
\EventShortTitle{}
\EventAcronym{}
\EventYear{}
\EventDate{}
\EventLocation{}
\EventLogo{}
\SeriesVolume{}
\ArticleNo{}

\hypersetup{
colorlinks=true, 
breaklinks=true, 
bookmarksnumbered,
pdftitle={}, 
pdfauthor={\textcopyright}, 
pdfsubject={}, 
pdfkeywords={}, 
pdfcreator={pdfLaTeX}, 
pdfproducer={LaTeX with hyperref and ClassicThesis} 
}

\usepackage[T1]{fontenc} 

\usepackage[utf8]{inputenc} 
\usepackage{fix-cm}

\usepackage{graphicx} 

\usepackage{lipsum} 

\usepackage{subcaption}

\usepackage{amsmath,amssymb,amsthm,latexsym} 

\usepackage{bm} 

\usepackage{semantic} 

\usepackage[shortcuts]{extdash} 

\usepackage{xspace} 

\usepackage{cellspace}
\usepackage{makecell}
\usepackage[dvipsnames]{xcolor}
\usepackage{fancyvrb}
\usepackage{tikz}
\usepackage{tikzscale}
\usepackage{pgfplots}
\usepackage{graphicx}
\usepackage[ruled,vlined]{algorithm2e}
\usepackage{hyperref}
\usepackage{tcolorbox}
\tcbuselibrary{listings,breakable}
\usepackage{mathtools}
\usepackage[inkscapelatex=false]{svg}

\usepackage{listings} 
\usepackage{upquote} 

\providecommand{\abs}[1]{\ensuremath{\left\lvert#1\right\rvert}}

\providecommand{\max}[1]{\ensuremath{\max\left(#1\right)}}
\providecommand{\min}[1]{\ensuremath{\min\left(#1\right)}}

\providecommand{\mset}[1]{\ensuremath{\left\{#1\right\}}}

\mathlig{|}{\mid}
\mathlig{==}{\equiv}
\mathlig{/==}{\not\equiv}
\mathlig{~>}{\leadsto}
\mathlig{~=}{\approx}
\mathlig{~}{\lnot}
\mathlig{|->}{\mapsto}
\mathlig{||-}{\Vdash}
\mathlig{|/=}{\nvDash}
\mathlig{|=}{\vDash}
\mathlig{|/-}{\nvdash}
\mathlig{?}{\mathbin{?}}
\mathlig{||}{\parallel}
\mathlig{<=>}{\Leftrightarrow}
\mathlig{<==>}{\Longleftrightarrow}

\newcommand{\ie}{i.e.\xspace}

\newcommand\extrafootertext[1]{%
    \bgroup
    \renewcommand\thefootnote{\fnsymbol{footnote}}%
    \renewcommand\thempfootnote{\fnsymbol{mpfootnote}}%
    \footnotetext[0]{#1}%
    \egroup
}

\definecolor{BashBG}{HTML}{0B0F14}
\definecolor{BashFG}{HTML}{E6E6E6}
\definecolor{BashAccent}{HTML}{7FDBCA}

\lstdefinestyle{bashstyle}{
  language=bash,
  basicstyle=\ttfamily\footnotesize,
  breaklines=true,
  columns=fullflexible,
  keepspaces=true,
  showstringspaces=false,
  upquote=true,
  keywordstyle=\color{NavyBlue}\bfseries,
  commentstyle=\color{OliveGreen},
  stringstyle=\color{BrickRed},
  backgroundcolor=\color{white}
}

\newtcblisting{bashblock}{
  listing engine=listings,
  listing only,
  breakable,
  colback=white,
  colframe=black,
  boxrule=0.5pt,
  arc=0pt,
  outer arc=0pt,
  left=0mm,
  right=0mm,
  top=0mm,
  bottom=0mm,
  boxsep=2mm,
  listing options={style=bashstyle}
}

\newtcolorbox{terminalbox}[1][]{
  colback=BashBG,coltext=BashFG,
  colframe=BashAccent, fonttitle=\bfseries,
  listing only, breakable, enhanced,
  left=1mm, right=1mm, top=1mm, bottom=1mm,
  listing options={language=bash,breaklines=true,basicstyle=\ttfamily\footnotesize},
  title=#1
}

\usetikzlibrary{arrows.meta}
\usetikzlibrary{arrows}
\usepgfplotslibrary{fillbetween}

\NewDocumentCommand{\FE}{o}{%
  \IfNoValueTF{#1}{%
    \texttt{FE}%
  }{%
    \ensuremath{\texttt{FE}_{#1}}
  }%
}

\newcommand{\afs}{\texttt{AFSAT}}
\pgfplotsset{compat=1.18}

\begin{document}
\date{\today}
\maketitle


\begin{abstract}
We study parallel Continuous Local Search (CLS) as a solution approach for Boolean satisfiability problems with symmetric pseudo-Boolean (PB) constraints.
Here, the $n$-variable PB-satisfiability problem is relaxed to a continuous optimisation problem with a differentiable objective function on an $n$-dimensional hypercube.
For satisfiable instances, the global minimisers of this optimisation problem correspond to satisfying assignments of the SAT problem at hand.
We present several novel findings via empirical experiments:
(i) redundant constraints can inhibit rather than accelerate convergence;
(ii) CLS shows promise as a sub-solver in hybridised settings, quickly completing partial assignments; and 
(iii) local search rapidly converges to a stable distribution of solution quality (\ie, degree of satisfaction), due to saddle-dense objectives where additional solver steps yield diminishing returns.
Our findings inform practical uses of CLS for SAT on modern accelerator hardware.
\end{abstract}


\section{Introduction}

Continuous local search (CLS) offers an alternative that is inherently parallelisable and amenable to GPU acceleration. CLS relaxes Boolean variables to real values on $[-1,1]$ via the Walsh-Fourier expansion~\cite{odonnell14}, reformulating satisfaction as bounded non-convex continuous optimisation. This approach was formalised in \texttt{FourierSAT}~\cite{kyrillidis2021solving} and extended for GPU computation in \texttt{FastFourierSAT}~\cite{cen2025massively}. A companion paper~\cite{christopher2026afsat} describes the engineering and performance of our solver Accelerated Fourier SAT (\afs{}), used throughout this work for empirical evaluation. We make the following contributions:
\begin{enumerate}
  \item We analyse the representational advantage of native psuedo-Boolean (PB) encodings for CLS over CNF translations (\S\ref{sec:rep_advantage}).
  \item We present closed-form Fourier coefficient derivations for all symmetric PB constraint types of practical interest (Appendix~\ref{sec:coefficients}).
  \item We provide empirical case studies revealing novel phenomena in CLS behaviour, including gradient interaction effects, partial assignment solving, and convergence profiling (\S\ref{sec:studies}).
\end{enumerate}


\section{Background}
\label{sec:background}

\subsection{Walsh-Fourier Transformation}
\label{sec:wft}

We review analytical properties of Boolean functions, notably that they can be relaxed into continuous functions in $\mathbb{R}$, with $\mset{\texttt{True},\texttt{False}}\mapsto\mset{-1,1}$. For a Boolean formula $\phi:\mset{\top, \bot}^{n} |-> \mset{\top,\bot}$, a continuous relaxation is $\text{Rel}_\phi: [-1, 1]^{n} |-> [-1, 1]$. $[-1,1]^n$ is the convex Boolean hypercube $\mathcal{Q}^n$. The Walsh-Fourier expansion~\cite{odonnell14, kyrillidis2021solving} is such a relaxation:
\begin{equation}
  \label{eqn:wft}
  \FE[\phi](\mathbf{X}) \stackrel{\text{def}}{=} \sum_{S\in 2^{\mathbf{X}}}\hat{f}(S)\prod_{x_i\in S} x_i
\end{equation}
\noindent $\mathbf{X}$ is the set of $n$ variables and $\hat{f}(S)$ are the \emph{Fourier coefficients} at subset $S \subseteq \mathbf{X}$. $\FE[\phi]$ is a multi-linear polynomial---a linear combination of \emph{elementary symmetric polynomials} (ESP).
\begin{equation}
  e_{i}^{n}=\sum_{1\le j_1<j_2<\ldots<j_i<n}{\prod_{k=1}^{i}{x_{j_k}}}
\end{equation}

Computing Fourier coefficients for general Boolean formulae may be exponentially expensive. However, for \emph{symmetric} constraints---those whose truth depends \emph{only on how many} literals are true---the symmetry reduces the number of distinct coefficients to $n+1$, and closed-form solutions exist~\cite{kyrillidis2021solving}. The general coefficient formula for symmetric constraints is:
\begin{align}
  \label{eqn:wfcoef}
  \hat{f}(S) =
  \begin{dcases}
    \frac{1}{2^{n}}\sum_{i=0}^{n}\binom{n}{i}sp_i & \abs{S} = 0 \\
    \frac{g(\rho)_{[\rho^{\abs{S}-1}]}}{\binom{n-1}{\abs{S} -1}2^{n-1}} & \abs{S} \neq 0
  \end{dcases}
\end{align}
where $sp_i = -1$ if $f$ is true when $i$ variables are true and $1$ otherwise; $g(\rho) = \sum_{i=0}^{n-1}d_i\binom{n-1}{i}(1+\rho)^{n-i-1}(1-\rho)^{i}$; and $d_i = \frac{1}{2}(sp_i - sp_{i+1})$ is the discrete derivative of the spectrum. We provide closed form solutions in Appendix~\ref{sec:coefficients}.

\subsection{Pseudo-Boolean Representation}

Conjunctive normal form (CNF) constrains each clause to be a disjunction of literals:
\begin{equation}
  \phi_{CNF} = \bigwedge_{C_k\in\phi}{\left(\bigvee_{l_{i}\in C}{l_i}\right)}
\end{equation}
Many problems are more naturally expressed using \emph{pseudo-Boolean} (PB) constraints of the form $\sum_{i=1}^{n}c_il_i\ \mathcal{R}\ k$, where $\mathcal{R}$ is a comparison operator, $c_i$ are coefficients, and $k$ is a threshold~\cite{barth95}. Important special cases---all symmetric with unitary coefficients---include at-most-one (\texttt{AMO}), exactly-one (\texttt{EO}), exactly-$k$ (\texttt{EK}) not-all-equal (\texttt{NAE}), exclusive-or (\texttt{XOR}), and cardinality-$k$ (\texttt{CARD}) constraints. Reduction of these constraints to CNF can be exponential na\"{i}vely, and polynomial encodings introduce auxiliary variables and clauses~\cite{sinz2005towards,bailleux2003efficient,warners1998linear,10.1007/978-3-031-95973-8_8}.

\subsection{Satisfaction as Optimisation}

For a formula $\phi$ decomposed into $m$ symmetric constraints $C_1,\ldots,C_m$ over variables $\mathbf{X}$, the sound sum-of-expansions yields a continuous objective~\cite{kyrillidis2021solving}:
\begin{equation}
  \min_{\mathbf{X}} \sum_{k=1}^{m}\FE[C_k](\mathbf{X}) \text{ subject to } \mathbf{X} \in \mathcal{Q}^n
\end{equation}
A satisfying assignment is achieved when $\sum_k \FE[C_k](\texttt{sign}(\mathbf{X})) = -m$. This non-convex, differentiable formulation motivates projected gradient descent (PGD), with the DFT enabling fully vectorised evaluation on GPU accelerators~\cite{cen2025massively}.

A remark on an important structural property: since every multi-linear polynomial is harmonic (trivial Laplacian), all critical points in the interior of $\mathcal{Q}^n$ are saddle points~\cite{kyrillidis2021solving}. Local and global minima exist only at the boundary, but the interior is proliferated with saddle points at which first-order methods like PGD can stall.

\subsection{Representational Advantage of PB Encodings}
\label{sec:rep_advantage}

The Walsh-Fourier expansion preserves the native variable count and constraint count of PB-encoded problems, in contrast to CNF translations which typically inflate both. In Appendix~\ref{sec:coefficients}), Table~\ref{tab:pbt} summarises the asymptotic costs.

The Fourier expansion maintains a representational advantage since no auxiliary variables or additional constraints are required; the dominant coefficient computation cost is incurred once and depends only on the constraint parameters $n$ and $k$.

\begin{table}[htb]
  \centering
  \begin{tabular}{|l|c|c|c|c|}
    \hline
    \textbf{Problem (Size)} & \multicolumn{2}{c|}{\textbf{CNF Translation}} & \multicolumn{2}{c|}{\textbf{Walsh-Fourier}} \\
    & Vars & Clauses & Vars & Constraints \\
    \hline
    Ramsey 3C, 13N & 2587 & 14183 & \textbf{\underline{234}} & \textbf{\underline{276}} \\
    Ramsey 3C, 16N & 4552 & 27032 & \textbf{\underline{360}} & \textbf{\underline{411}} \\
    Costas Array 5x5 & 97 & 590 & 97 & \textbf{\underline{245}} \\
    Costas Array 10x10 & 712 & 10622 & 712 & \textbf{\underline{3434}} \\
    Costas Array 15x15 & 2577 & 61980 & 2577 & \textbf{\underline{18825}} \\
    Costas Array 20x20 & 5872 & 196367 & 5872 & \textbf{\underline{57289}} \\
    Sudoku 9x9 & 729 & 3257 & 729 & \textbf{\underline{341}}\\
    \hline
  \end{tabular}
  \caption{Variable and clause counts: Walsh-Fourier vs. CNF encodings for problems with native PB formulations. Bold underlined values indicate outright smaller counts.}
  \label{tab:pbcompact}
\end{table}

Table~\ref{tab:pbcompact} illustrates this advantage on concrete problem instances. For Ramsey colouring problems with \texttt{EO} constraints, the Walsh-Fourier encoding reduces both variable and clause counts by an order of magnitude over CNF. For Costas array problems, variable counts are preserved while clause counts are substantially reduced.

This representational advantage is significant beyond mere compactness. Resolution proof systems---and thereby CDCL procedures---can find PB constraints encoded in CNF challenging~\cite{10.1145/7531.8928}, while systematic search tailored to PB constraints should theoretically be stronger~\cite{COOK198725}. Practical realisation of this advantage in systematic search, however, remains difficult~\cite{10.1007/978-3-319-94144-8_18,10.1007/978-3-319-94144-8_5}. CLS provides an alternative avenue for exploiting native PB encodings on accelerator hardware.


\section{Case Studies}
\label{sec:studies}

We present three case studies that reveal distinct aspects of CLS behaviour. All experiments use \afs{} on NVIDIA Tesla V100 GPUs; see the companion tool paper~\cite{christopher2026afsat} for full experimental setup details.

\subsection{Ramsey Colouring: Gradient Interactions and Encoding Effects}
\label{sec:ramsey}

A Ramsey colouring of a complete graph $G=(V,E)$ with colours $C$ is an edge colouring containing no monochromatic triangles, amongst other conditions~\cite{Kowalski2015,Alm2016401AB,Alm2019}. Whether colourings exist for $|C|\in \{8,13\}$ remains open.

We study a \emph{balanced assignment} relaxation requiring each vertex to have equal edge-colour distribution. In PB-SAT:
\begin{align*}
  \bigwedge
  \begin{dcases}
    \forall v \in V, &\;\;\sum_{i\in C}{c(v,i)}=1\\
    \forall v \in V, \forall i \in C, &\sum_{v'\in V\setminus v}{c(v,i)}\geq \frac{\abs{V}-1}{\abs{C}}
  \end{dcases}
\end{align*}
Here the \texttt{EO} constraint ensures each edge gets exactly one colour, and cardinality constraints enforce balancing. With this encoding, \afs{} finds colourings near-instantly for $(|V|,|C|)=(16,3)$.
We note that $(\abs{V},\abs{C})=(16,4)$ reaches constraint lengths close to the practical floating-point ceiling ($n\approx 50$), while global cardinality constraints on edges (120 literals for $(16,3)$) are well beyond it, necessitating per-vertex decomposition.

\subsubsection{Redundant Constraint Effect}
An alternative encoding decomposes the \texttt{EO} constraint as a disjunction plus a cardinality over negated literals: $\texttt{CARD}_{\geq 2}(\lnot\mathbf{x})\texttt{+OR}(\mathbf{x})$. We combine both encodings---\texttt{EO} and the decomposition---in a single problem. Since $\texttt{CARD}_{\geq 2}\texttt{+OR} = \texttt{EO} - 1$, the combination is equivalent to $2\cdot\texttt{EO}-1$, doubling the gradient magnitude of the edge-colouring constraints.

\begin{figure}[htb]
  \centering
  \includegraphics[width=0.95\linewidth]{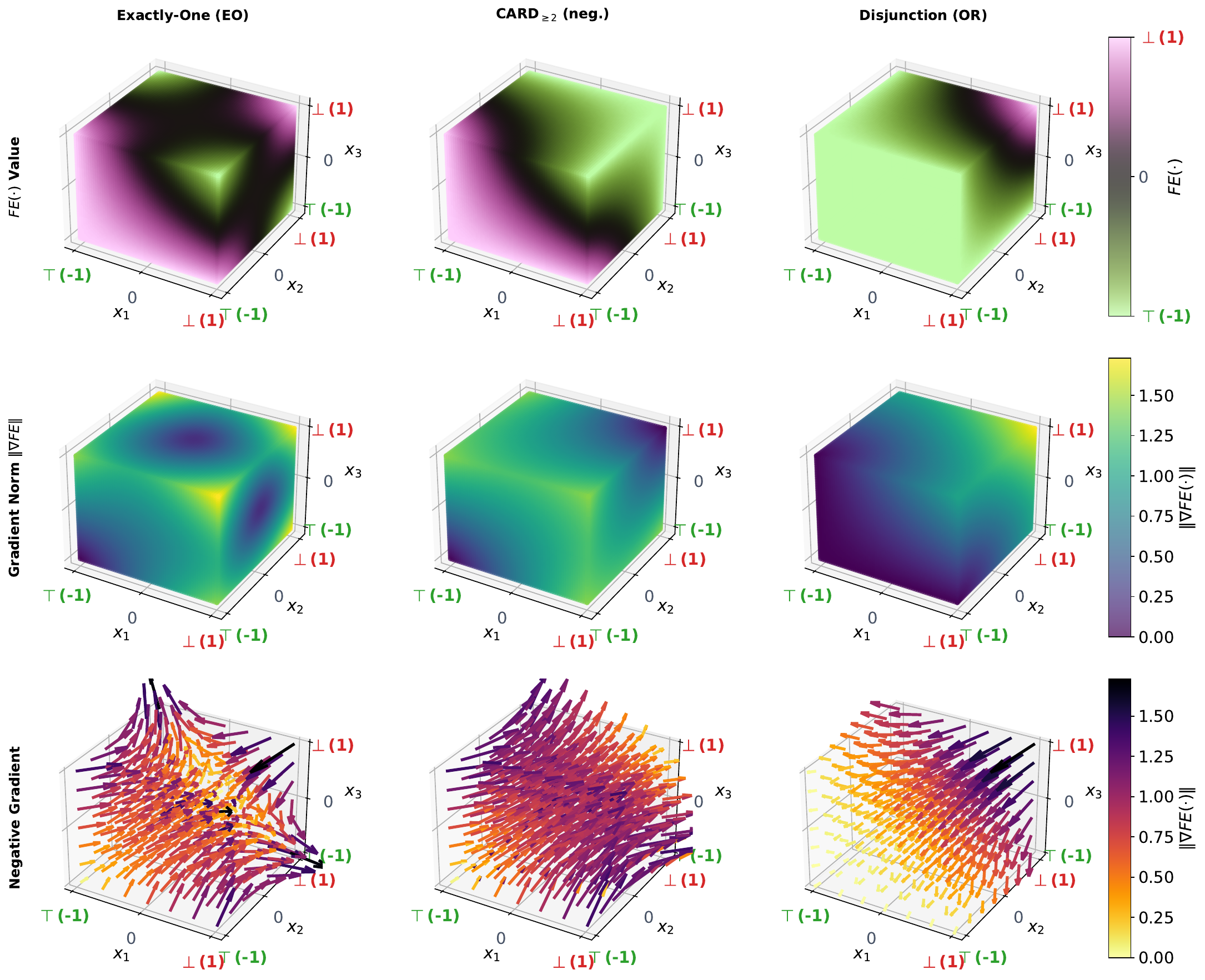}
  \caption{Evaluation, slope, and gradients of EO, $\text{CARD}_{\geq2}(\lnot\mathbf{x})$, and OR constraints in 3 variables.}
  \label{fig:combpb}
\end{figure}

Counterintuitively, this doubled gradient steepness \emph{inhibits} convergence: CLS succeeds approximately once per hundred runs (vs. near-instant success otherwise). Failed solutions typically violate 1--3 balancing constraints while satisfying all edge-colouring constraints.

We hypothesise a damped-oscillator mechanism:
\begin{enumerate}
  \item Satisfying assignments occupy a tiny fraction of the hypercube vertices. Edge-colouring constraints are satisfied at specific corners (Figure~\ref{fig:combpb}).
  \item To satisfy remaining balancing constraints, literals must be flipped, requiring traversal away from the currently occupied corner.
  \item The gradient field near satisfied vertices of the edge-colouring constraints strongly attracts back toward the corner. With doubled encoding, this attractive force is doubled.
  \item The opposing gradient from unsatisfied balancing constraints is overwhelmed, causing the assignment to oscillate with decreasing amplitude and converge at the current (non-satisfying) position.
\end{enumerate}

\begin{figure}[htb]
  \centering
  \includegraphics[width=0.95\linewidth]{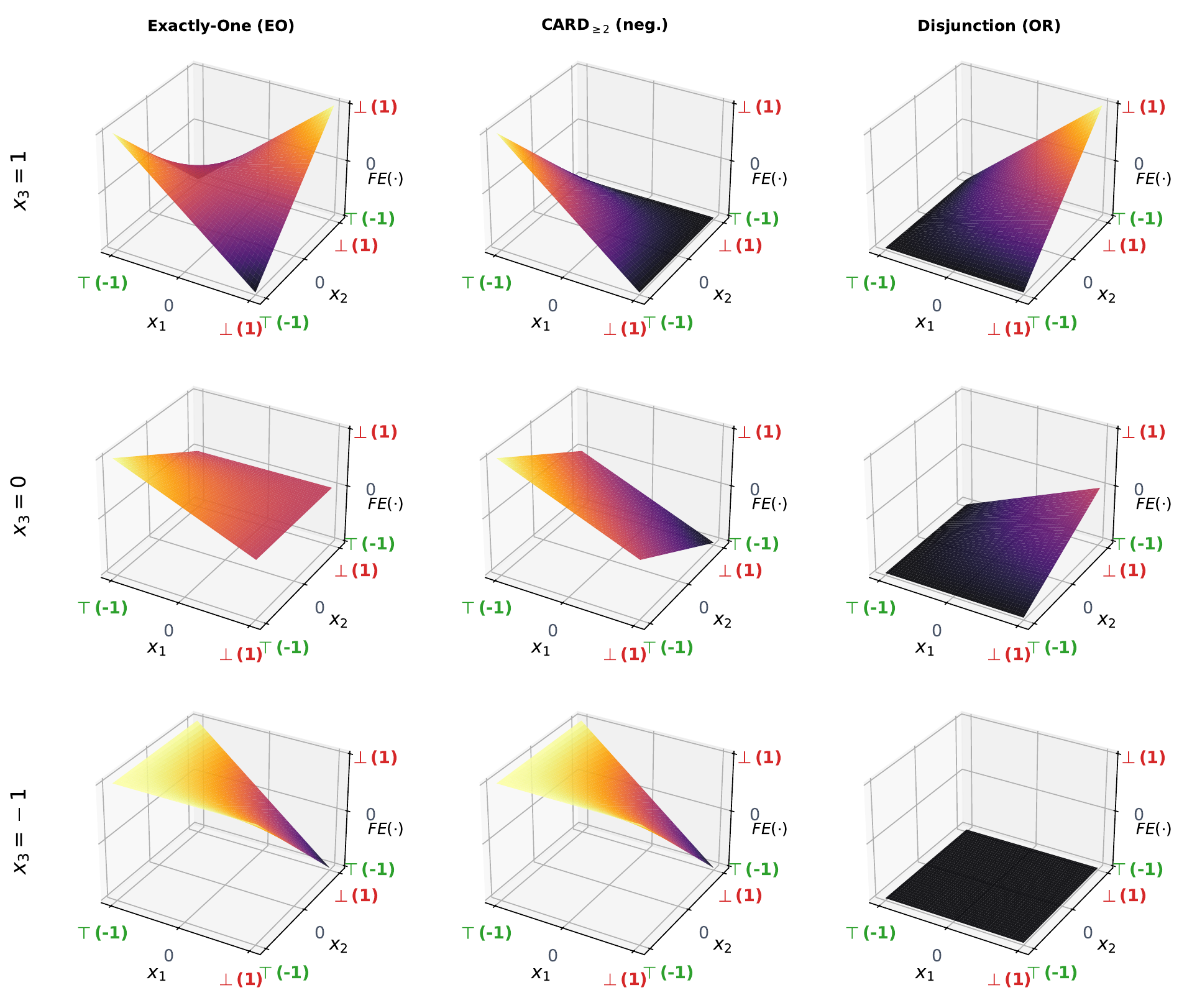}
  \caption{3D surface of evaluations for \texttt{EO}, $\text{\texttt{CARD}}_{\geq2}(\lnot\mathbf{x})$, and \texttt{OR} in 3 variables with one variable fixed at $\pm1$ and $0$.}
  \label{fig:combpb2d}
\end{figure}

We further observe (Figure~\ref{fig:combpb2d}) that the \texttt{EO} constraint exhibits a plateau at the all-true corner (all literals $=-1$), where the gradient magnitude is small. This creates a region from which PGD may struggle to escape, as the small step size satisfies convergence criteria prematurely.

These findings suggest that:
\begin{itemize}
  \item CLS benefits from \emph{minimal} encodings, consistent with the representational advantage of native PB formulations.
  \item Adaptive constraint weighting---informed by per-constraint gradient magnitudes---is a promising direction for improving CLS on problems with structural symmetry.
\end{itemize}

\subsection{Hard Random 3-SAT: Partial Assignment Completion}
\label{sec:hard_rand}

We evaluate CLS on hard random satisfiable 3-SAT instances from the 2007 SAT Competition,\footnote{\url{https://satisfiability.org/competition/2007/}} with 450--650 variables at the clause-to-variable phase transition ratio of $\approx 4.26$~\cite{Cheeseman:1991:WRH,doi:10.1126/science.1073287}. These instances are solved rapidly by \textsc{Dagster}~\cite{10.1007/978-3-031-20862-1_6}, which incorporates \textsc{gNovelty+}~\cite{DBLP:journals/jsat/PhamTGS08}.

CLS alone does not solve any of these instances. To determine what support is needed, we fix a partial assignment of a known satisfying assignment and allow CLS to complete the remainder, simulating integration with a systematic solver.

\begin{figure}[!htbp]
  \centering
  \includegraphics[width=\linewidth]{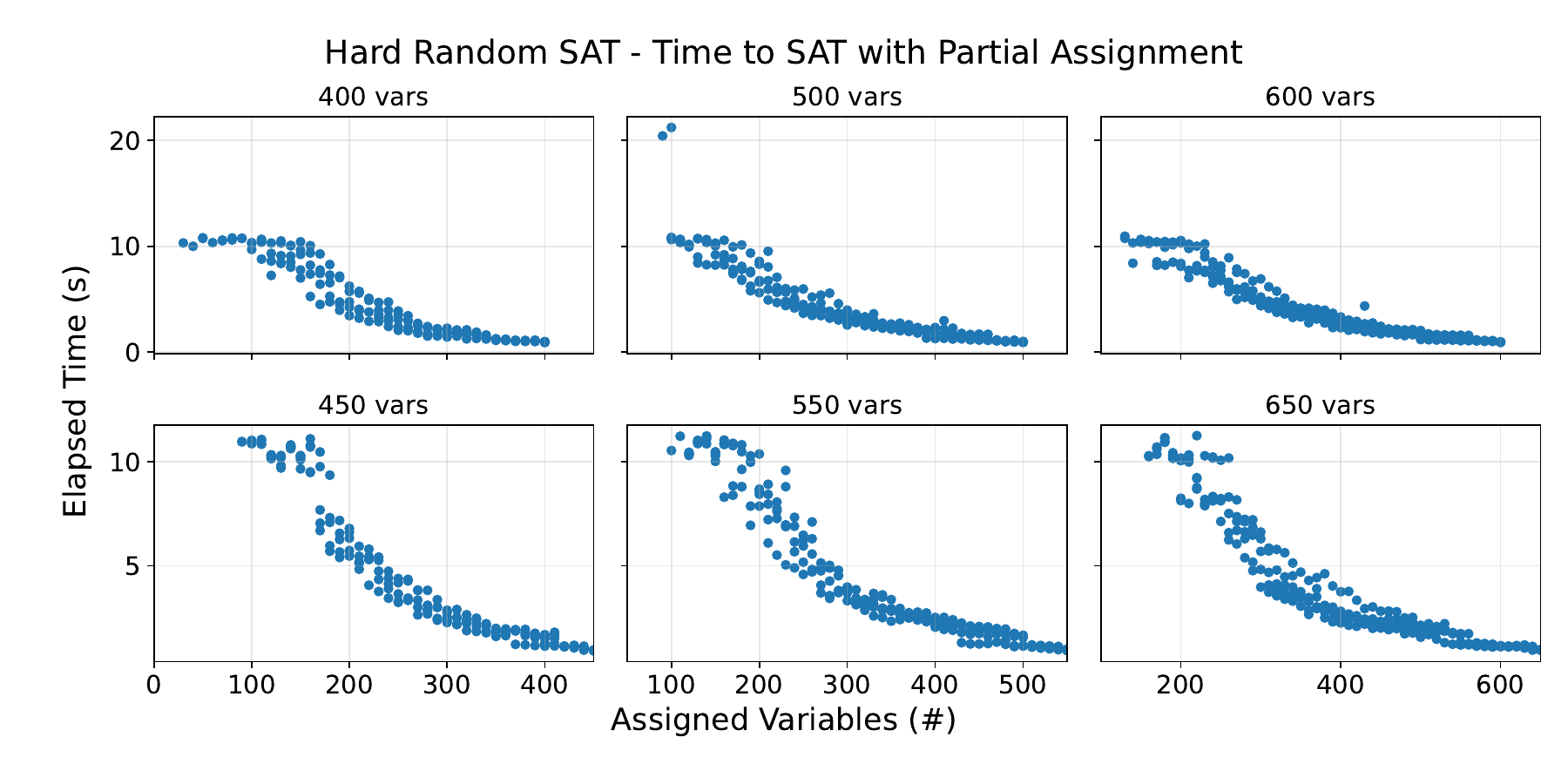}
  \caption{Time to solution for hard random 3-SAT instances (400--650 variables) with increasing partial assignment lengths.}
  \label{fig:prog_hard_rand}
\end{figure}

Figure~\ref{fig:prog_hard_rand} shows that \afs{} can complete solutions once $15$--$20\%$ of variables are fixed, with monotonically decreasing solution time as the partial assignment lengthens. This on-average monotonic behaviour is significant: it means intuitions about when and how to deploy CLS as a sub-solver can be relied upon, opening the door to integrated architectures where systematic search provides partial assignments that CLS completes on accelerators. This is particularly relevant for \#SAT procedures that must enumerate completions of partial assignments.

\subsection{Costas Arrays: Convergence Profile Analysis}
\label{sec:costas}
A Costas array is a set of $n$ points in an $n \times n$ grid with exactly one point per row and column, and all displacement vectors distinct~\cite{10.1007/978-3-031-20862-1_6}. Whether arrays exist at $n \in \{32, 33\}$ are open problems~\cite{5464772}.
For $n\leq 6$, \afs{} finds all Costas solutions (with duplication). For $n\geq 7$, solution-finding drops substantially; beyond $n\geq 8$, none are found without assistance. 

We discover distinct convergence profiles that depend on the fraction of variables assigned. These profiles have direct implications for solver parameterisation---specifically, the maximum descent steps hyper-parameter.

\begin{figure}[htbp]
  \centering
    \includegraphics[width=\linewidth]{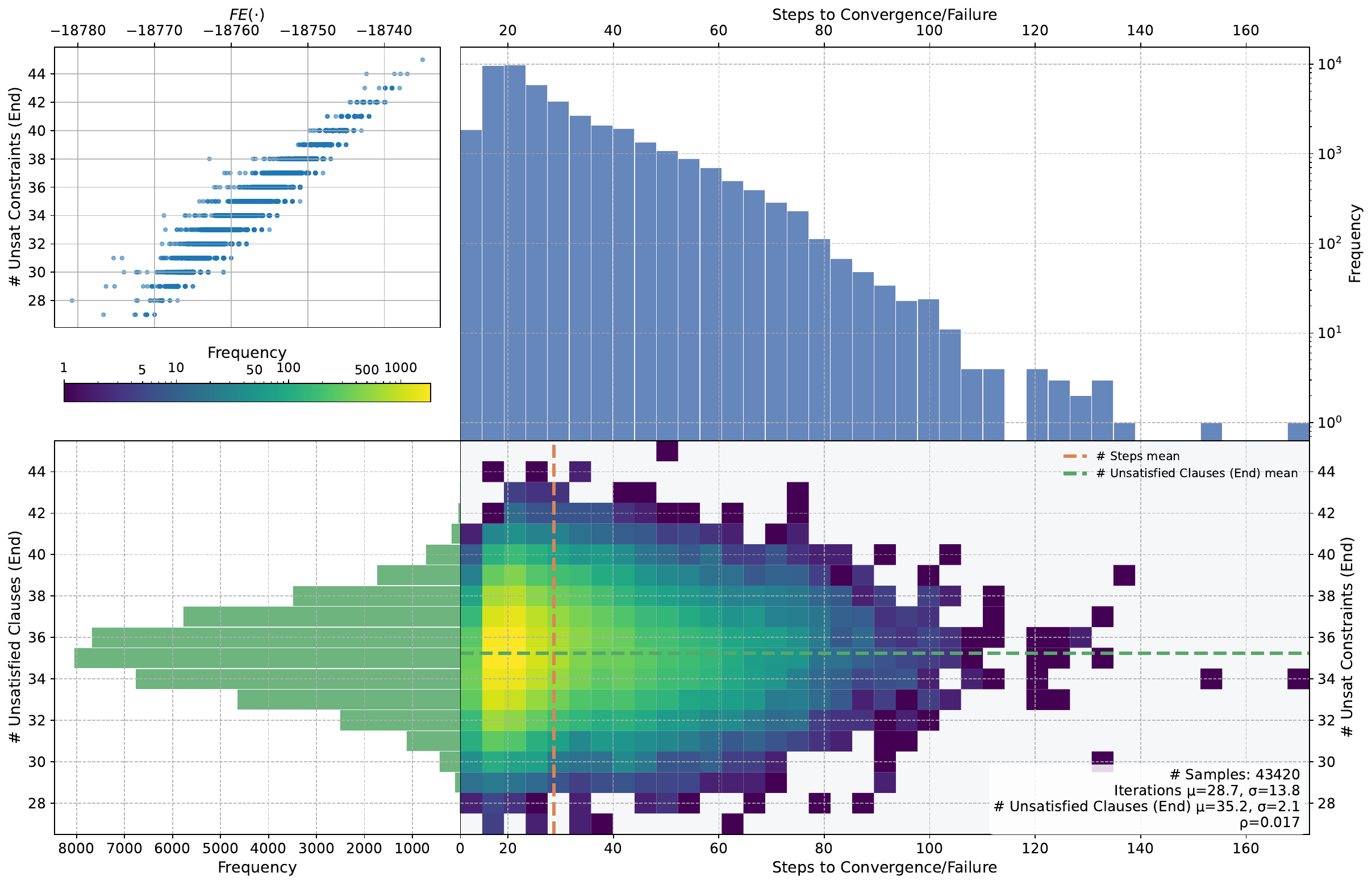}
    \caption{Costas-15 ($\approx$66\% assigned): We show a heatmap view of a 3D histogram, and the component 2D histograms flanking above (iterations taken to converge) and left (number of constraints unsatisfied). The top-right quadrant shows the correlation between the count of unsatisfied constraints and the energy of the evaluation.}
    \label{fig:costas15-flat}
  \end{figure}

\begin{figure}[htbp]
\centering
\includegraphics[width=\linewidth]{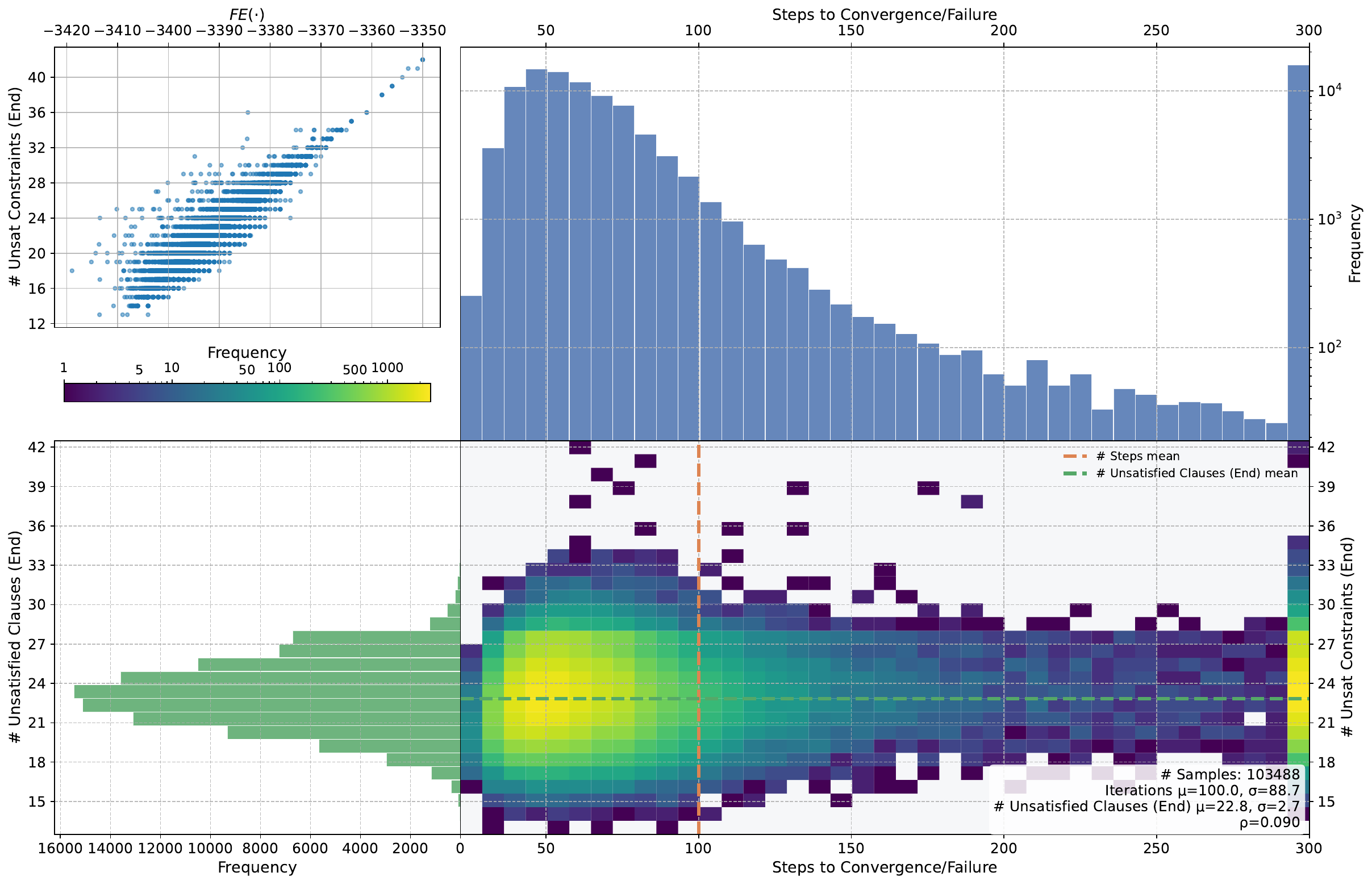}
\caption{Costas-10 ($\approx$10\% assigned): We show the same view for this problem as in Figure~\ref{fig:costas15-flat}}
\label{fig:costas10-flat}
\end{figure}

For a high partially assigned regime ($\approx$66\% assigned), Figure~\ref{fig:costas15-flat} shows a unimodal distribution where nearly all starting assignments converge well before the maximum step limit. Solution quality is normally distributed and uncorrelated with the number of steps taken. Comparing this to a low partially assigned regime ($\approx$10\% assigned), Figure~\ref{fig:costas10-flat} shows a qualitatively different profile: a substantial fraction of initial assignments reach the maximum step limit (300) without converging. The quality of unconverged assignments is statistically indistinguishable from converged ones, and convergence, when achieved, occurs early.

This difference has important implications. In vectorised CLS, all batch elements must iterate together; idling converged elements waste compute. The convergence profile directly determines the cost of the maximum-steps hyperparameter. For unimodal profiles, setting the limit near the distribution's central mass minimises waste. For bimodal profiles, this becomes a throughput-quality trade-off: a lower limit improves throughput but terminates unconverged descents, while a higher limit wastes cycles on already-converged elements.

Note that Kyrillidis et al.~\cite{kyrillidis2021solving} prove a polynomial upper bound on the number of PGD steps for convergence of order $O(|\text{\#vars}||\text{\#constraints}|)$, which greatly exceeds the empirically observed step counts. Furthermore, this bound assumes algebraic precision, which cannot be maintained on accelerator hardware with limited floating-point precision.


\section{Related Work}

The CLS approach builds on the Fourier analysis of Boolean functions~\cite{odonnell14} and its application to SAT~\cite{kyrillidis2021solving}. GPU-accelerated evaluation via DFT was introduced by Cen et al.~\cite{cen2025massively}. The representational advantages of PB constraints over CNF have been studied extensively~\cite{sinz2005towards,bailleux2003efficient,warners1998linear,10.1007/978-3-031-95973-8_8}, with proof-complexity results showing that resolution (and CDCL) can struggle with PB constraints in CNF~\cite{10.1145/7531.8928,COOK198725,10.1007/978-3-319-94144-8_18,10.1007/978-3-319-94144-8_5}. State-of-the-art PB solvers such as \textsc{SCIP}~\cite{10.5555/3215359.3215361,BolusaniEtal2024OO} use mixed-integer programming techniques but remain inherently sequential. Parallel SAT solving has been studied extensively~\cite{DBLP:journals/aim/HamadiW13,katsirelos:2013}, but GPU-based approaches remain non-competitive with state-of-the-art sequential systems. 
The CLS approach builds on the Fourier analysis of Boolean functions~\cite{odonnell14} and its application to SAT~\cite{kyrillidis2021solving}. GPU-accelerated evaluation via DFT was introduced by Cen et al.~\cite{cen2025massively}. The representational advantages of PB constraints over CNF have been studied extensively~\cite{sinz2005towards,bailleux2003efficient,warners1998linear,10.1007/978-3-031-95973-8_8}, with proof-complexity results showing that resolution (and CDCL) can struggle with PB constraints in CNF~\cite{10.1145/7531.8928,COOK198725,10.1007/978-3-319-94144-8_18,10.1007/978-3-319-94144-8_5}. State-of-the-art PB solvers such as \textsc{SCIP}~\cite{10.5555/3215359.3215361,BolusaniEtal2024OO} use mixed-integer programming techniques but remain inherently sequential. Parallel SAT solving has been studied extensively~\cite{DBLP:journals/aim/HamadiW13,katsirelos:2013}, but GPU-based approaches remain non-competitive with state-of-the-art sequential systems.


\section{Conclusions and Future Work}
We studied continuous local search in GPU accelerated computing environments.
We catalogued a family of symmetric pseudo-Boolean constraints that offer a representational advantage over CNF.
In our setting the compactness of pseudo-Boolean problem representation enables a higher number of parallel local searches.
Our empirical evaluation has revealed several novel phenomena:
\begin{itemize}
  \item Problems that are expressed with redundant constraints cause search to exhibit gradient imbalances that inhibit convergence, suggesting that minimal PB encodings are preferable for CLS.
  \item CLS can reliably complete partial assignments for hard random $3$-SAT with $15$--$20\%$ of variables fixed, with monotonically improving performance as more variables are fixed, thus has potential for use as a sub-solver.
  \item Convergence profiles exhibit characteristic distributional shapes that depend on problem structure and the fraction of free variables, with direct implications for parameter selection.
\end{itemize}

Future work will pursue: {\em (i)} principled adaptive weighting schemes for heterogeneous PB problems, informed by per-constraint gradient magnitude analysis; {\em (ii)} integration of CLS into distributed search architectures as exemplified by \textsc{Dagster}; {\em (iii)} automated identification of PB substructures in CNF formulae and thereby compilation of CNF problems to compact representation amenable to CLS; and {\em (iv)} investigation of CLS for discovering problem constitutionality and dependency structures.


\bibliography{bib}

@misc{christopher2026afsat,
  title={Accelerated Fourier SAT (AFSAT): Fully Realising a GPU-based Symmetric Pseudo-Boolean SAT Solver}, 
  author={Cody J Christopher and Charles Gretton},
  year={2026},
  eprint={2606.06641},
  archivePrefix={arXiv},
  primaryClass={cs.AI},
  url={https://arxiv.org/abs/2606.06641}, 
}

@book{odonnell14, 
    place={Cambridge}, 
    title={Analysis of Boolean Functions}, 
    publisher={Cambridge University Press}, 
    author={O'Donnell, Ryan}, 
    year={2014}
}

@article{cen2025massively,
  title={Massively parallel continuous local search for hybrid SAT solving on GPUs},
  author={Cen, Yunuo and Zhang, Zhiwei and Fong, Xuanyao},
  journal={Proceedings of the AAAI Conference on Artificial Intelligence},
  volume={39},
  number={11},
  pages={11140--11149},
  year={2025}
}

@article{kyrillidis2021solving,
  title={Solving hybrid Boolean constraints in continuous space via multilinear Fourier expansions},
  author={Kyrillidis, Anastasios and Shrivastava, Anshumali and Vardi, Moshe Y and Zhang, Zhiwei},
  journal={Artificial Intelligence},
  volume={299},
  pages={103559},
  year={2021},
  publisher={Elsevier}
}

@inproceedings{sinz2005towards,
  title={Towards an optimal CNF encoding of boolean cardinality constraints},
  author={Sinz, Carsten},
  booktitle={International conference on principles and practice of constraint programming},
  pages={827--831},
  year={2005},
  organization={Springer}
}

@inproceedings{bailleux2003efficient,
  title={Efficient CNF encoding of boolean cardinality constraints},
  author={Bailleux, Olivier and Boufkhad, Yacine},
  booktitle={International conference on principles and practice of constraint programming},
  pages={108--122},
  year={2003},
  organization={Springer}
}

@article{warners1998linear,
  title={A linear-time transformation of linear inequalities into conjunctive normal form},
  author={Warners, Joost P},
  journal={Information Processing Letters},
  volume={68},
  number={2},
  pages={63--69},
  year={1998},
  publisher={Elsevier}
}

@inproceedings{10.1007/978-3-031-95973-8_8,
author="Bierlee, Hendrik
and Dekker, Jip J.
and Stuckey, Peter J.",
editor="Tack, Guido",
title="Revisiting Pseudo-Boolean Encodings from an Integer Perspective",
booktitle="Integration of Constraint Programming, Artificial Intelligence, and Operations Research",
year="2025",
publisher="Springer Nature Switzerland",
address="Cham",
pages="113--133",
isbn="978-3-031-95973-8"
}

@inproceedings{10.1007/978-3-319-94144-8_5,
author = {Elffers, Jan and Gir\'{a}ldez-Cru, Jes\'{u}s and Nordstr\"{o}m, Jakob and Vinyals, Marc},
title = {Using Combinatorial Benchmarks to Probe the Reasoning Power of Pseudo-Boolean Solvers},
year = {2018},
isbn = {978-3-319-94143-1},
publisher = {Springer-Verlag},
address = {Berlin, Heidelberg},
doi = {10.1007/978-3-319-94144-8_5},
booktitle = {Theory and Applications of Satisfiability Testing – SAT 2018: 21st International Conference, SAT 2018, Held as Part of the Federated Logic Conference, FloC 2018, Oxford, UK, July 9–12, 2018, Proceedings},
pages = {75–93},
numpages = {19},
location = {Oxford, United Kingdom}
}

@techreport{BolusaniEtal2024OO,
  author = {Suresh Bolusani and Mathieu Besan{\c{c}}on and Ksenia Bestuzheva and Antonia Chmiela and Jo{\~{a}}o Dion{\'{i}}sio and Tim Donkiewicz and Jasper van Doornmalen and Leon Eifler and Mohammed Ghannam and Ambros Gleixner and Christoph Graczyk and Katrin Halbig and Ivo Hedtke and Alexander Hoen and Christopher Hojny and Rolf van der Hulst and Dominik Kamp and Thorsten Koch and Kevin Kofler and Jurgen Lentz and Julian Manns and Gioni Mexi and Erik~M\"{u}hmer and Marc E. Pfetsch and Franziska Schl{\"o}sser and Felipe Serrano and Yuji Shinano and Mark Turner and Stefan Vigerske and Dieter Weninger and Lixing Xu},
  title = {{The SCIP Optimization Suite 9.0}},
  type = {Technical Report},
  institution = {Optimization Online},
  month = {February},
  year = {2024},
  url = {https://optimization-online.org/2024/02/the-scip-optimization-suite-9-0/}
}

@article{10.5555/3215359.3215361,
author = {Shinano, Yuji and Heinz, Stefan and Vigerske, Stefan and Winkler, Michael},
title = {FiberSCIP-A Shared Memory Parallelization of SCIP},
year = {2018},
issue_date = {February 2018},
publisher = {INFORMS},
address = {Linthicum, MD, USA},
volume = {30},
number = {1},
issn = {1526-5528},
journal = {INFORMS J. on Computing},
month = feb,
pages = {11–30},
numpages = {20}
}

@article{Alm2016401AB,
  title={401 and beyond: improved bounds and algorithms for the Ramsey algebra search},
  author={Jeremy F. Alm},
  journal={ArXiv},
  year={2016},
  volume={abs/1609.01817},
  url={https://api.semanticscholar.org/CorpusID:910341}
}

@article{Kowalski2015,
  author    = {Tomasz Kowalski},
  title     = {Representability of Ramsey Relation Algebras},
  journal   = {Algebra Universalis},
  year      = {2015},
  volume    = {74},
  number    = {3},
  pages     = {265--275},
  doi       = {10.1007/s00012-015-0353-0},
  issn      = {1420-8911}
}

@article{Alm2019,
  author    = {Jeremy F. Alm and David A. Andrews},
  title     = {A reduced upper bound for an edge-coloring problem from relation algebra},
  journal   = {Algebra Universalis},
  year      = {2019},
  volume    = {80},
  number    = {2},
  pages     = {19},
  doi       = {10.1007/s00012-019-0592-6},
  issn      = {1420-8911}
}

@inproceedings{10.1007/978-3-319-94144-8_18,
author = {Vinyals, Marc and Elffers, Jan and Gir\'{a}ldez-Cru, Jes\'{u}s and Gocht, Stephan and Nordstr\"{o}m, Jakob},
title = {In Between Resolution and Cutting Planes: A Study of Proof Systems for Pseudo-Boolean SAT Solving},
year = {2018},
isbn = {978-3-319-94143-1},
publisher = {Springer-Verlag},
address = {Berlin, Heidelberg},
doi = {10.1007/978-3-319-94144-8_18},
booktitle = {Theory and Applications of Satisfiability Testing – SAT 2018: 21st International Conference, SAT 2018, Held as Part of the Federated Logic Conference, FloC 2018, Oxford, UK, July 9–12, 2018, Proceedings},
pages = {292–310},
numpages = {19},
location = {Oxford, United Kingdom}
}

@article{COOK198725,
title = {On the complexity of cutting-plane proofs},
journal = {Discrete Applied Mathematics},
volume = {18},
number = {1},
pages = {25-38},
year = {1987},
issn = {0166-218X},
doi = {10.1016/0166-218X(87)90039-4},
author = {W. Cook and C.R. Coullard and Gy. Turán}
}

@article{10.1145/7531.8928,
author = {Urquhart, Alasdair},
title = {Hard examples for resolution},
year = {1987},
issue_date = {Jan. 1987},
publisher = {Association for Computing Machinery},
address = {New York, NY, USA},
volume = {34},
number = {1},
issn = {0004-5411},
doi = {10.1145/7531.8928},
journal = {J. ACM},
month = jan,
pages = {209–219},
numpages = {11}
}

@misc{barth95,
  title={A Davis-Putnam based enumeration algorithm for linear pseudo-Boolean optimization},
  author={Barth, Peter},
  year={1995},
  publisher={Max-Planck-Institut f{\"u}r Informatik}
}

@article{DBLP:journals/aim/HamadiW13,
  author    = {Youssef Hamadi and
               Christoph M. Wintersteiger},
  title     = {Seven Challenges in Parallel {SAT} Solving},
  journal   = {{AI} Mag.},
  volume    = {34},
  number    = {2},
  pages     = {99--106},
  year      = {2013},
  doi       = {10.1609/aimag.v34i2.2450},
  bibsource = {dblp computer science bibliography, https://dblp.org}
}

@article{DBLP:journals/jsat/PhamTGS08,
  author    = {Duc Nghia Pham and
               John Thornton and
               Charles Gretton and
               Abdul Sattar},
  title     = {Combining Adaptive and Dynamic Local Search for Satisfiability},
  journal   = {J. Satisf. Boolean Model. Comput.},
  volume    = {4},
  number    = {2-4},
  pages     = {149--172},
  year      = {2008},
  doi       = {10.3233/sat190042},
  bibsource = {dblp computer science bibliography, https://dblp.org}
}

@inproceedings{Cheeseman:1991:WRH,
  author       = {Cheeseman, Peter and Kanefsky, Bob and Taylor, William M.},
  title        = {Where the Really Hard Problems Are},
  booktitle    = {Proceedings of the 12th International Joint Conference on Artificial Intelligence (IJCAI)},
  year         = {1991},
  pages        = {331--337},
  address      = {Sydney, Australia},
  publisher    = {IJCAI},
}

@article{
doi:10.1126/science.1073287,
author = {M. Mézard  and G. Parisi  and R. Zecchina },
title = {Analytic and Algorithmic Solution of Random Satisfiability Problems},
journal = {Science},
volume = {297},
number = {5582},
pages = {812-815},
year = {2002},
doi = {10.1126/science.1073287},
eprint = {https://www.science.org/doi/pdf/10.1126/science.1073287}}

@inproceedings{katsirelos:2013,
	author = {Katsirelos, George and Sabharwal, Ashish and Samulowitz, Horst and Simon, Laurent},
	title = {Resolution and Parallelizability: Barriers to the Efficient Parallelization of {SAT} Solvers},
	booktitle = {Proceedings of the Twenty-Seventh {AAAI} Conference on Artificial Intelligence},
	series = {AAAI'13},
	year = {2013},
	url = {http://www.aaai.org/ocs/index.php/AAAI/AAAI13/paper/viewFile/6421/7193}
}

@INPROCEEDINGS{5464772,
  author={Russo, Jon C. and Erickson, Keith G. and Beard, James K.},
  booktitle={2010 44th Annual Conference on Information Sciences and Systems (CISS)}, 
  title={Costas array search technique that maximizes backtrack and symmetry exploitation}, 
  year={2010},
  volume={},
  number={},
  pages={1-8},
  doi={10.1109/CISS.2010.5464772}}

@InProceedings{10.1007/978-3-031-20862-1_6,
author="Burgess, Mark Alexander
and Gretton, Charles
and Milthorpe, Josh
and Croak, Luke
and Willingham, Thomas
and Tiu, Alwen",
editor="Khanna, Sankalp
and Cao, Jian
and Bai, Quan
and Xu, Guandong",
title="Dagster: Parallel Structured Search with Case Studies",
booktitle="PRICAI 2022: Trends in Artificial Intelligence",
year="2022",
publisher="Springer Nature Switzerland",
address="Cham",
pages="75--89",
isbn="978-3-031-20862-1"
}


\appendix
\section{Closed-Form Fourier Coefficients for Symmetric PB Constraints}
\label{sec:coefficients}

We present closed-form solutions for the Fourier coefficients of all symmetric PB constraint types supported by CLS. In each case, $S$ denotes a subset of variables, with $|S|$ substitutable for $S$ due to symmetry, and $n$ is the total number of variables.

\textbf{Disjunction (\texttt{OR}):}
\begin{align*}
  \widehat{\texttt{OR}}(S) =
  \begin{dcases}
    \frac{1}{2^{n-1}} - 1 &\abs{S} = 0 \\
    \frac{1}{2^{n-1}} &\abs{S} \neq 0
  \end{dcases}
\end{align*}

\textbf{Exactly-One (\texttt{EO}):}
\begin{align*}
  \widehat{\texttt{EO}}(S) =
  \begin{dcases}
    1-\frac{n}{2^{n-1}} & \abs{S} = 0 \\
    \frac{2\abs{S}-n}{2^{n-1}} & \abs{S} \neq 0
  \end{dcases}
\end{align*}

\textbf{Exactly-$k$ (\texttt{EK}):}
\begin{align*}
    \widehat{{\texttt{E}_{k}}}(S) &=
        \begin{dcases}
            1-\frac{\binom{n}{k}}{2^{n-1}} & \abs{S} = 0 \\
            \frac{g_{\texttt{EK}}(\rho)_{[\rho^{|S|-1}]}}
                 {\binom{n-1}{|S|-1}2^{n-1}} & \abs{S} \neq 0
        \end{dcases}\\
    g_{\texttt{EK}}(\rho) &= \frac{((2k-n) + n\rho)}{k}\binom{n-1}{k-1}(1+\rho)^{n-k-1}(1-p)^{k-1}\\
    &=\frac{1}{n}\binom{n}{k}((2k-n) + n\rho)(1+\rho)^{n-k-1}(1-p)^{k-1}
\end{align*}

\textbf{At-Most-One (\texttt{AMO}):}
\begin{align*}
  \widehat{\texttt{AMO}}(S) =
  \begin{dcases}
    1-\frac{n-1}{2^{n-1}} & \abs{S} = 0 \\
    \frac{2\abs{S}-n-1}{2^{n-1}} & \abs{S} \neq 0
  \end{dcases}
\end{align*}

\textbf{Not-All-Equal (\texttt{NAE}):}\\
The spectrum $sp_{\texttt{NAE}}=\mset{1,-1,-1,\ldots,-1,1}$ simplifies the noise operator to $g(\rho)=(1+\rho)^{n-1} - (1-\rho)^{n-1}$. By the binomial theorem, even-order terms cancel:
\begin{align*}
  \widehat{\texttt{NAE}}(S) =
  \begin{dcases}
    \frac{1}{2 ^ {n - 2}}-1 & \abs{S} = 0 \\
    \frac{1}{2 ^ {n - 2}} & \abs{S}\ \ \text{even}\\
    0 & \abs{S}\ \ \text{odd}
  \end{dcases}
\end{align*}

\textbf{Exclusive-Or (\texttt{XOR}):} \\
The alternating spectrum $sp_{\texttt{XOR}}=\mset{1,-1,1,-1,\ldots}$ produces total cancellation of all terms in $g(\rho)$ except $2^{n-1}\rho^{n-1}$. Thus \texttt{XOR} has a single non-zero coefficient:
\begin{align*}
  \widehat{\texttt{XOR}}(S) =
  \begin{dcases}
    0 & \abs{S} \neq [n] \\
    1 & \abs{S} = [n]
  \end{dcases}
\end{align*}

\textbf{Cardinality-$k$ (\texttt{CARD}):}
\begin{align*}
  &\widehat{{\texttt{CARD}}}_{\ge k}(S) =
  \begin{dcases}
    1-\frac{\sum_{i=k}^{n}\binom{n}{i}}{2^{n-1}} & \abs{S} = 0 \\
    \frac{\binom{n-1}{k-1}\left(g_{\texttt{CARD}}(\rho)\right)_{[\rho^{\abs{S}-1}]}}{\binom{n-1}{\abs{S}-1}2^{n-1}} & \abs{S} \neq 0
  \end{dcases}\\
  &g_{\texttt{CARD}}(\rho) = (1+\rho)^{n-k}(1-\rho)^{k-1}
\end{align*}

These closed-form solutions enable polynomial-time coefficient computation for all constraint types, providing the foundation for CLS on heterogeneous PB problems.
\clearpage
\section{CNF Encoding Tables}
\begin{table}[htb]
  \centering
  \resizebox{\textwidth}{!}{%
    \begin{tabular}{|Sl|Sl|Sl|Sl|}
      \hline
      \textbf{Clause Type $\phi$} & \textbf{PB Form} & \textbf{$\FE[\phi]$ / $O(\hat{f}(S))$} & \textbf{$O(\#\texttt{CNF}(f))$} \\
      \hline
      Disjunction (\texttt{OR}) & $\sum_nx_n \ge 1$ & $O(1)$ & --- \\
      \hline
      At most one (\texttt{AMO}) & $\sum_nx_n \le 1$ & $O(n)$ & Na\"{i}ve $O(n^2)$, $O(\log n)$ \\
      \hline
      Exactly one (\texttt{EO}) & $\sum_nx_n = 1$ & $O(n)$ & Na\"{i}ve $O(n^2)$, $O(\log n)$ \\
      \hline
      Exactly $k$ (\texttt{EK}) & $\sum_nx_n = k$ & $O(n\log^{2}{n})$ & See Table~\ref{tab:cardCNF}.\\
      \hline
      Not all equal (\texttt{NAE}) & $\bigwedge
      \begin{cases}\textstyle\sum_nx_n < n\\\textstyle\sum_nx_n > 0
      \end{cases}$ & $O(1)$ & $O(1)$ \\
      \hline
      Exclusive Or (\texttt{XOR}) & $\sum_nx_n \equiv 1 \pmod{2}$ & $O(n)$ & $O(n)$ \\
      \hline
      Cardinality-$k$ (\texttt{CARD}) & $\sum_nx_n \ge k$ & $O(n\log^{2}{n})$ & See Table~\ref{tab:cardCNF}.\\
      \hline
  \end{tabular}}
  \caption{Asymptotic costs of Fourier coefficient computation vs. CNF translation for symmetric PB constraints.}
  \label{tab:pbt}
\end{table}
\begin{table}[htb]
  \centering
  \begin{tabular}{|Sl|Sl|Sl|Sl|}
    \hline
    \textbf{Encoding} & \textbf{$\#$clauses} & \textbf{$\#$aux. vars}\\
    \hline
    Na\"ive & $\binom{n}{k+1}$ & $0$\\
    \hline
    Sequential unary counter & $O(n \cdot k)$ & $O(n \cdot k)$\\
    \hline
    Parallel binary counter & $7n - 3\lfloor\log n\rfloor - 6$ & $2n - 2$\\
    \hline
    Bailleux and Boufkhad~\cite{bailleux2003efficient} & $O(n^2)$ & $O(n \cdot \log n)$\\
    \hline
    Warners~\cite{warners1998linear} & $8n$ & $2n$\\
    \hline
  \end{tabular}
  \caption{CNF encodings for Cardinality-$k$~\cite{sinz2005towards}. All non-na\"{i}ve encodings introduce auxiliary variables.}
  \label{tab:cardCNF}
\end{table}
\end{document}